# A MODULARITY COMPARISON OF LONG SHORT-TERM MEMORY AND MORPHOGNOSIS NEURAL NETWORKS

Thomas E. Portegys, portegys@gmail.com , ORCID 0000-0003-0087-6363
Kishwaukee College, Malta, Illinois USA

## ABSTRACT

This study compares the modularity performance of two artificial neural network architectures: a Long Short-Term Memory (LSTM) recurrent network, and Morphognosis, a neural network based on a hierarchy of spatial and temporal contexts. Mazes are used to measure performance, defined as the ability to utilize independently learned mazes to solve mazes composed of them. A maze is a sequence of rooms connected by doors. The modular task is implemented as follows: at the beginning of the maze, an initial door choice forms a context that must be retained until the end of an intervening maze, where the same door must be chosen again to reach the goal. For testing, the door-association mazes and separately trained intervening mazes are presented together for the first time. While both neural networks perform well during training, the testing performance of Morphognosis is significantly better than LSTM on this modular task.

## 1. INTRODUCTION

Modularity is an important organization principle in artificial systems (Baldwin & Clark, 2000). Briefly, modularity is the ability to combine reusable components of learned behavior to solve an overarching task rather than learning the task from beginning to end every time a portion of it changes. Both humans and animals are adept at this (Clune et al., 2013; Lorenz et al., 2011). Modularity allows a task to be composed of a sequence or hierarchy of subtasks, where subtasks can be switched in and out in response to environmental variations while maintaining competency on an overall task (Portegys, 2015). For artificial neural networks (ANNs) modular learning is a step toward the long-sought goals of achieving a balance between plasticity and stability.

Maze navigation has been a staple of measuring the learning capabilities of animals for decades (Carr, 1913; Miles, 1930; Olton, 1979). For example, the effects of a proposed memory-enhancing drug may be tested on rats by observing how well they learn mazes in comparison to a control group. Mazes have also been employed to investigate learning in artificial neural networks (ANNs) (Blynel & Floreano, 2003; Johansson & Lansner, 2002; Yamauchi & Beer, 1995). These studies typically use variations of a T-maze, where the learner is faced with multiple intersections each having two choices of direction leading toward a goal. Learning can be accomplished either through exploration or by training, the method used here.

This study is a follow-up to a previous comparison of the Mona ANN with Elman and LSTM networks using a maze-learning task (Portegys, 2010). Mona is a goal-seeking system that learns hierarchies of cause-and-effect relationships. However, it lacks the non-linear function approximation capability that multi-layer perceptrons, such as LSTM and Morphognosis, possess.

A maze is defined as a sequence of rooms connected by doors. The modular task is implemented as follows: at the beginning of the maze, an initial door choice forms a context that must be retained until the end of an intervening maze, where the same door must be chosen again to reach the goal. For training, door-associations with and without intervening mazes are learned. Separately trained mazes are also learned as maze modules. For testing, maze modules are incorporated as intervening mazes between door-associations for the first time. Success is achieved if a modular maze is successfully navigated and the door-association context is retained such that the correct door is chosen at the end of the task.

A comparison is done between two artificial neural network architectures: a Long Short-Term Memory (LSTM) network, and Morphognosis, a neural network based on a hierarchy of spatial and temporal contexts.

The Long Short-Term Memory (LSTM) network (Hochreiter & Schmidhuber, 1997) is designed to allow state information to be retained over extended periods of time using special activation units called memory blocks. LSTM networks have been shown to solve tasks that require long-term retention of state information (Gers et al., 2000).

The Morphognosis model has been used to simulate pufferfish nest-building (Portegys, 2019), and honey bee foraging (Portegys, 2020), tasks that require both temporal and spatial processing.

Azam's (2000) survey of modular neural networks cites the Hierarchical Mixture of Experts (HME) as a prominent architecture. In this and several other schemes, an a priori partitioning of the input space is done to create a tree structure of neural networks. Tani and Nolfi (1999) devised a means of autonomously partitioning the input space of a maze learning task by recognizing a dynamical phase change as an indicator of a transition to a different partition. Optimally, each partition is then managed by a dedicated recurrent neural network. This system was able to learn a two-room maze but instabilities requiring further study were also observed.

In general, modular learning has been most successful when using explicitly designed separate networks to learn subtasks (Chang et al., 2019; Kirsch et al., 2018). The goal of ANNs self-modularizing, as biological networks do, appears elusive (Csordás et al., 2020). For this reason, the success of Morphognosis on the modular learning task is significant.

This next section first describes the maze implementation in detail. A brief description of the Morphognosis model follows that.

## 2. Description

### 2.1. Mazes

Mazes pose three learning tasks:

1. Maze learning: learn to navigate a sequence of rooms connected by doors.
2. Context learning: learn correspondences between room configurations separated by intervening mazes.
3. Modular learning: the intervening mazes are trained independently and presenting only during testing. This measures the ability to dynamically combine independently learned sequences to solve a maze.

A maze consists of a sequence of rooms connected by doors. There are a fixed number of doors. The learner outputs a door choice or a wait. A room contains a room-specific set of on/off marks.

There are four types of rooms:

1. Context begin room: in this room the learner is presented with marks having a single *on* value that corresponds to the correct door choice. This door leads to either a maze entry or directly to a context end room. If it leads to a maze entry room, an intervening maze must be navigated before reaching the context end room. In the context end room, the learner must choose the door that was marked *on* in the context begin room.
2. Maze entry room: the marks uniquely identify the configuration of the upcoming maze sequence, consisting of maze interior rooms. The learner uses this information to navigate these rooms.
3. Maze interior room: the maze entry room determine the correct door choice sequence to move through the maze.
4. Context end room: in this room the correct door choice is determined by the context begin room marks.

There are two types of mazes: a context maze, shown in Figure 1, containing context begin and end rooms and intervening maze rooms, and an independent maze, shown in Figure 2.

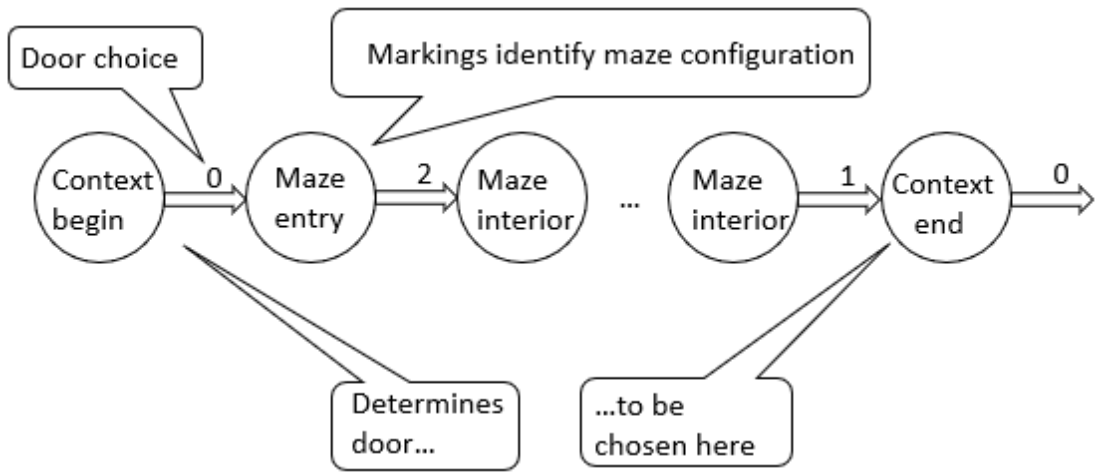

Figure 1 – Context maze.

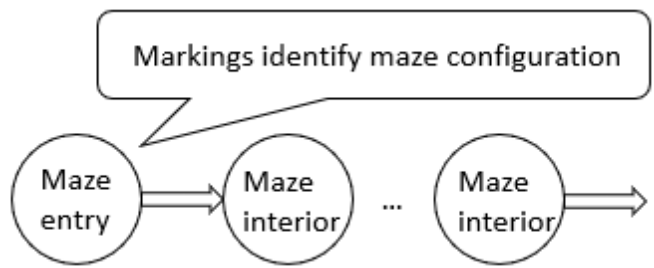

Figure 2 – Independent maze.

Learner input format:

<context room marks><maze entry room marks><maze interior room mark><context end room marks>

Output: A door choice or a wait response. These are encoded as a binary vector.

Here is an example of a context maze sequence:

context begin room: input = **[1 0 0]** [0 0 0 0 0 0] [0] [0 0 0]  output = *[1 0 0 0] (door 0)*

maze begin room:    input = [0 0 0] **[0 0 0 1 0 0]** [0] [0 0 0]  output = [0 0 1 0] (door 2)

maze interior room:  input = [0 0 0] [0 0 0 0 0 0] **[1]** [0 0 0]  output = [0 0 1 0] (door 2)

maze interior room:  input = [0 0 0] [0 0 0 0 0 0] **[1]** [0 0 0]  output = [1 0 0 0] (door 0)

maze interior room:  input = [0 0 0] [0 0 0 0 0 0] **[1]** [0 0 0]  output = [0 0 1 0] (door 2)

context end room:    input = [0 0 0] [0 0 0 0 0 0] [0] **[1 1 1]**  output = *[1 0 0 0] (door 0)*

There are three types of generated training sequences:

1. Door associations: context begin room and context end room.
2. Context mazes.
3. Independent mazes.

For testing, context-mazes are created by combining door associations with independent mazes embedded as intervening mazes.

## 2.2. MORPHOGNOSIS

Morphognosis is designed to be a general method of capturing contextual information that can enhance the power of an artificial neural network (ANN). It provides a framework for organizing spatial and temporal sensory events mapped to motor responses into a tractable format suitable for ANN training and usage.

Morphognosis has been used to model the locomotion and foraging of the C. elegans nematode worm (Portegys, 2018), the nest-building behavior of a pufferfish (Portegys, 2019), and honey bee foraging (Portegys, 2020).

For the maze task, two simplifications are made to the customizable Morphognosis model:

1. Since a maze are purely temporal task, spatial processing is not required.
2. Since input information for each time step is vital to navigate a maze, aggregation of inputs is not required,

As a result of these simplifications, the model closely resembles a Time Delay Neural Network (TDNN) (Waibel et al. 1989), in which the input is a sliding window of time steps.

Dual cooperating networks are used. One network tracked the entire time step sequence of inputs through a maze, and the other operated only on the current time step. The network producing the greatest response value was chosen as the output. This arrangement can be viewed as a subsumption architecture (Brooks, 1986), where lower-level processing can be overridden by higher levels. Note that although multiple networks are employed, both networks are trained on the same input sequences, thus there is no external partitioning of the input into separate networks.

As an example of how the two networks work together, navigating through the intervening maze requires a longer temporal context; however, choosing the correct door in the context begin room, and making an initial door choice in the maze entry room require only knowledge of the current input.

Morphognosis uses the MultiLayerPerceptron class in the Weka 3.8.3 machine learning library (https://www.cs.waikato.ac.nz/ml/weka/).

The Java and Python code is available on GitHub: https://github.com/morphognosis/Maze

## 3. RESULTS

These parameters were set for the Weka multilayer perceptron:

- a single hidden layer of 50 neurons
- learning rate = 0.1
- momentum = 0.2
- training epochs = 5000

The LSTM network used is in the Keras 2.2.4 machine learning package: https://keras.io/, with these parameters:

- a single hidden layer of 128 neurons
- mean squared error loss function
- adam optimizer
- training epochs = 1000

Three features were selected to train and test on:

1. Number of doors.
2. Length of the intervening maze.

3. Number of intervening context and independent mazes.

DOOR TRAINING AND TESTING

Holding the intervening maze length at 5 and the number of intervening context and independent mazes at 10 each, the number of doors were varied. Every door must be trained with every intervening context maze, meaning that each additional/fewer door raises/lowers the number of training samples by the number of intervening context mazes.

Figure 3 shows that training validation results and Figure 4 shows the testing results. A run is counted as correct if all doors are correctly chosen. 25 trials with different randomly generated mazes were averaged.

For training, the LSTM shows perfect performance, while the performance of Morphognosis is good. For testing, Morphognosis significantly outperforms LSTM.

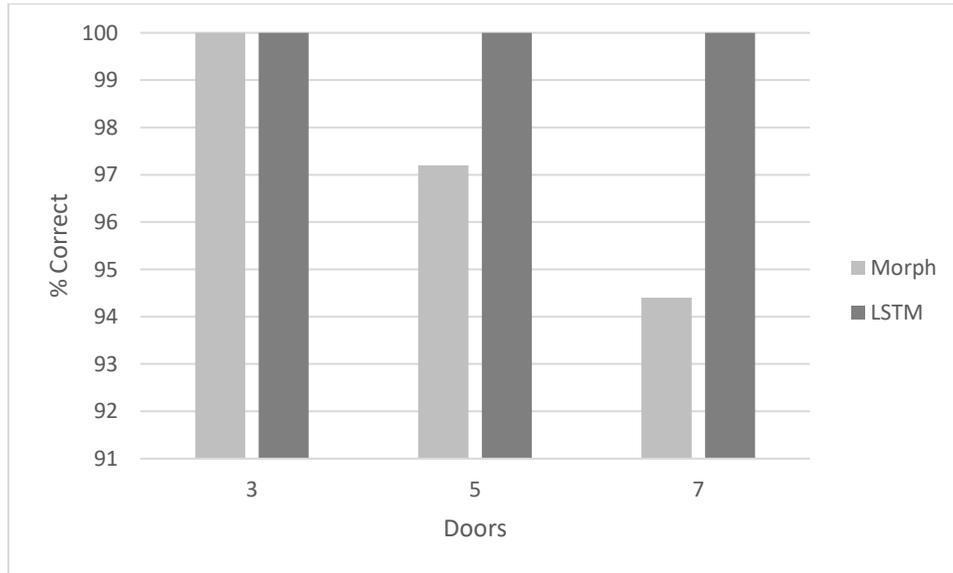

Figure 3 - Door training

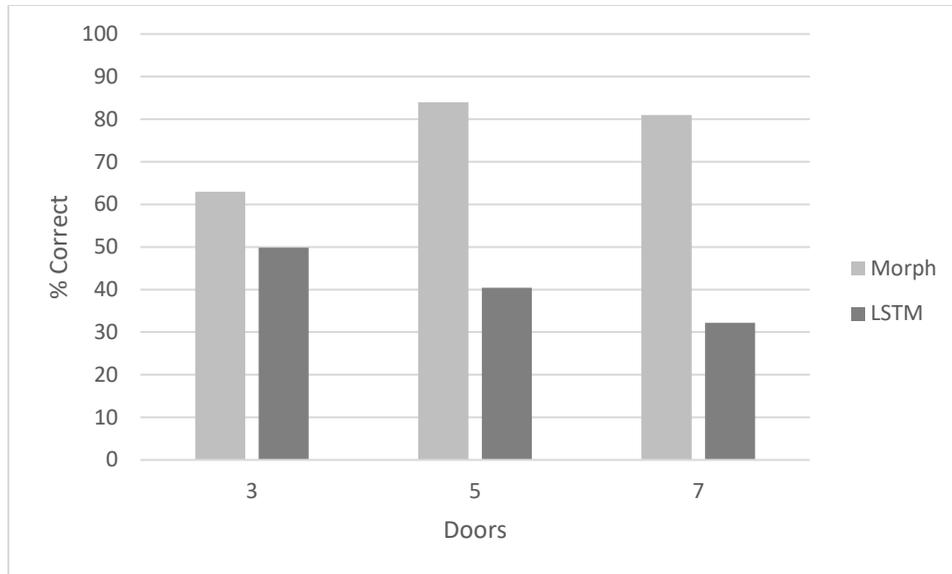

Figure 4 – Door testing

INTERVENING CONTEXT AND INDEPENDENT MAZES QUANTITY TRAINING AND TESTING

Holding the number of doors at 5, and the intervening maze length at 5, the number of intervening context mazes and independent mazes were varied. As with varying the number of doors, each additional/fewer intervening context maze raises/lowers the number of training samples by the number doors.

For training, shown in Figure 5, the LSTM again shows perfect performance, while the performance of Morphognosis is good. For testing, shown in Figure 6, Morphognosis significantly outperforms LSTM.

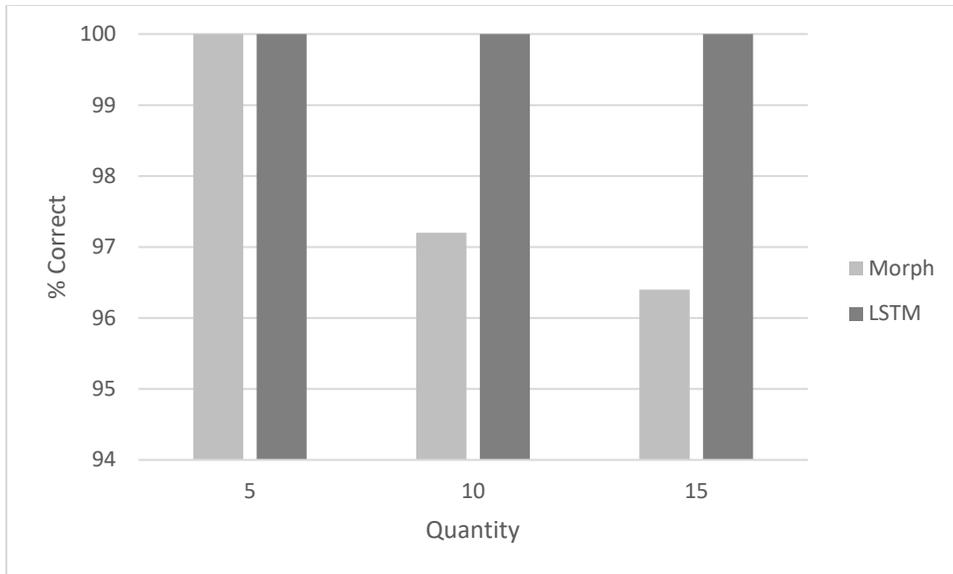

Figure 5 – Intervening context and independent maze quantity training

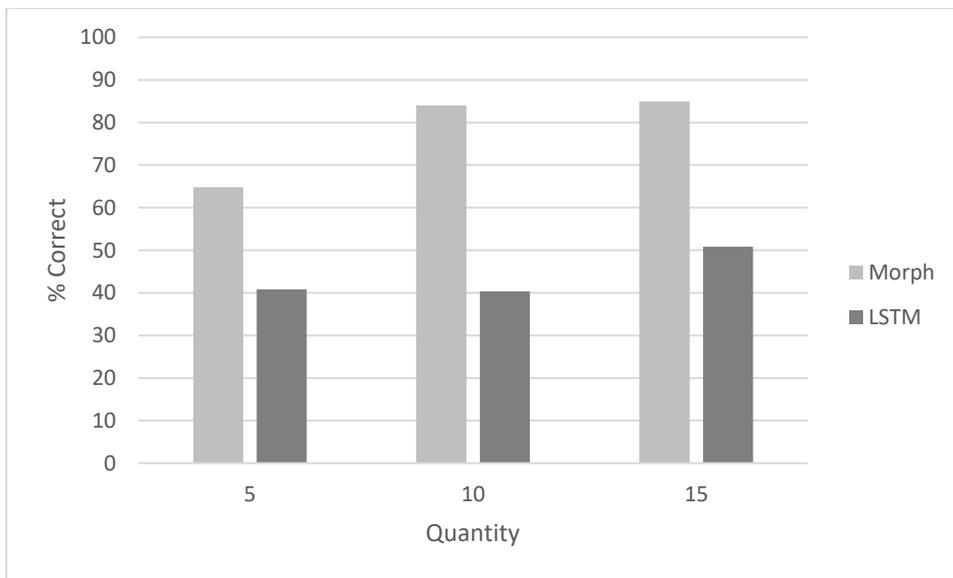

Figure 6 – Intervening context and independent maze quantity testing

INTERVENING MAZE LENGTH TRAINING AND TESTING

Holding the number of doors at 5, the number of intervening context mazes and independent mazes at 10, the length of the intervening mazes was varied. The length of the intervening maze poses a challenge to maintain temporal memory of the maze markings in the maze entry room which indicate the proper door sequence for the interior rooms in the intervening maze.

For training, shown in Figure 7, the LSTM again shows perfect performance, while the performance of Morphognosis is good. For testing, shown in Figure 8, Morphognosis significantly outperforms LSTM, although both fall off as the maze length increases.

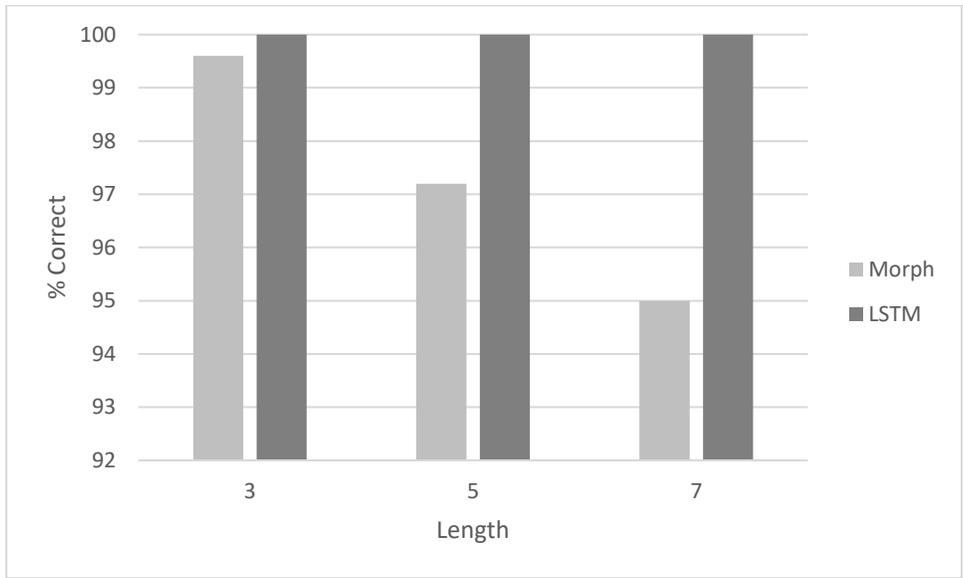

Figure 7 – Intervening maze length training

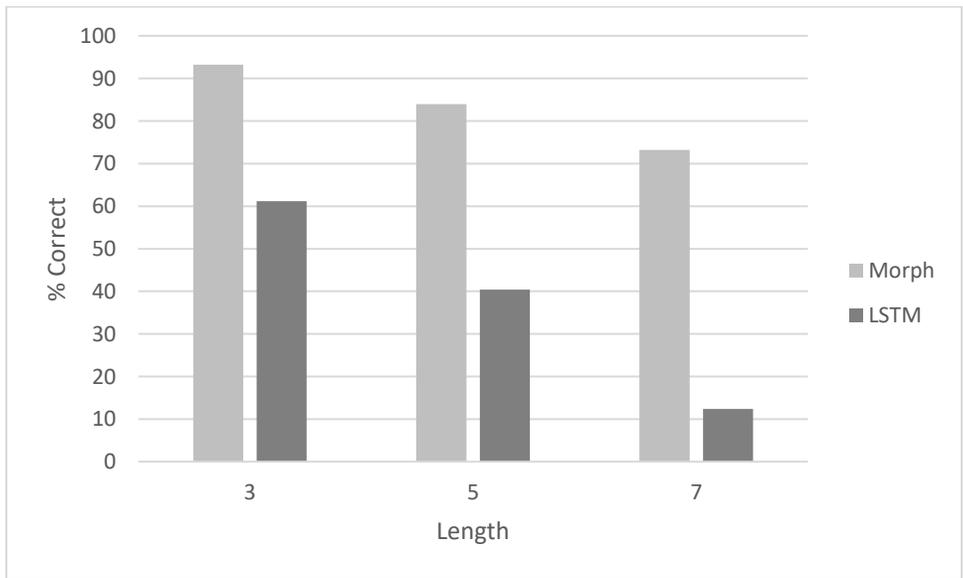

Figure 8 - Intervening maze length testing

## 4. DISCUSSION

The relative success of Morphognosis compared with LSTM on the maze-learning task can be attributed to two factors:

1. Context information is modularly recorded in the input rather than non-modularly entangled in the network. For example, context begin room information is available upon entering the context end room, allowing the correct door choice to be made independently of the intervening maze configuration. A key feature of Morphognosis is that sensory information coming from distant reaches of time and space is recorded and supplied to the network as distinguishable, and thus modular, input. This in turn allows modular processing.
2. The use of a subsumption architecture, featuring dual networks that learn from short-term and long-term time spans, allows the system to respond to either more local or more global contexts provided by the environment. For example, when a maze entry room configuration is encountered that has not previously been trained together with a preceding maze context room configuration, the local network can determine the correct response, since it has been trained to learn only from the current sensory input. In general, subsumption supports a stack of machines where a lower-level machine produces the system's output unless a higher-level machine, processing a more global context, overrides it with a better output.

The ability to learn and express modular behaviors that can be composed into larger structures remains a vexing problem for artificial neural network architectures. The poor performance of the LSTM network with modular training highlights the problem of using this type of neural network on tasks having a dynamic nature that may cause components to change, necessitating retraining. This is a serious issue since so many real-world environments pose challenges that fall into this category. For these tasks it would be beneficial to be able to retrain only the portions that change and have interactions with the rest of the system remain functional. Morphognosis is an effort in this direction.

## 5. FUTURE WORK

Capturing higher level contexts more effectively is a challenging goal for Morphognosis. Contexts serve to guide behavior over long temporal paths or within hierarchical structures. In both cases modular learning is of great value. To date implementations of Morphognosis have taken the form of a single vertical stack of context vectors, where each layer contains a single, possibly aggregated, vector. Taking a cue from unsupervised learning systems such as Adaptive Resonance Theory (ART) (Carpenter & Grossberg, 2003), storing multiple centroid vectors within a layer would retain multiple contexts. Each centroid would serve as a branching point in a tree of contexts, where each branch trains a separate neural network.

Combining reinforcement learning (Francois-Lavet et al. 2018) and modularity is a strength of the Mona goal-seeking neural network (Portegys, 2001). Mona accomplishes this through cause-and-effect learning, building in the process hierarchies of networks that serve as conduits for goal values motivating responses toward goals. Finding a way to implement cause-and-effect learning in Morphognosis, a multilayer perceptron system, would combine goal-seeking with the advantages of non-linear learning.